# A Hybrid Deep Learning Model for Arabic Text Recognition

Mohammad Fasha[1]

Amman, Jordan

Bassam Hammo[2], Nadim Obeid[3]

King Abdullah II School for Information Technology
The University of Jordan
Amman, Jordan

Jabir AlWidian[4]

Princess Sumayah School for Technology
Amman, Jordan

*Abstract*—Arabic text recognition is a challenging task because of the cursive nature of Arabic writing system, its joint writing scheme, the large number of ligatures and many other challenges. Deep Learning (DL) models achieved significant progress in numerous domains including computer vision and sequence modelling. This paper presents a model that can recognize Arabic text that was printed using multiple font types including fonts that mimic Arabic handwritten scripts. The proposed model employs a hybrid DL network that can recognize Arabic printed text without the need for character segmentation. The model was tested on a custom dataset comprised of over two million word samples that were generated using (18) different Arabic font types. The objective of the testing process was to assess the model's capability in recognizing a varied set of Arabic fonts representing a varied cursive styles. The model achieved good results in recognizing characters and words and it also achieved promising results in recognizing characters when it was tested on unseen data. The prepared model, the custom datasets and the toolkit for generating similar datasets are made publically available, these tools can be used to prepare models for recognizing other font types as well as to further extend and enhance the performance of the proposed model.

## I. INTRODUCTION

Optical Character Recognition (OCR) is the process of recognizing text in images and transforming it into a machine encoded text. OCR is an important research area and generally, it can be classified into two main groups, online OCR and offline OCR. Online OCR involves recognizing text while typing in real time such as recognizing digital stylus writing on mobile phones, while offline OCR involves the recognition of text in document images such as scanned documents archives, printed application forms, bank cheques, postal mail and many others. In addition, OCR addresses two main categories of text; machine printed text and handwritten text and each of these two areas has its own challenges. Printed text is faced with the challenge of the diverse font types and the various formatting styles as well as the quality of the printed and the scanned images, while handwritten text is considered more challenging because of the diverse writing styles of individuals.

Recognizing Arabic text in images has additional challenges that are mainly caused by the cursive nature of Arabic script. In addition, Arabic characters are connected in words, and the writing system has a large number of ligatures, which increase the challenge of recognizing text based on characters' features. Further, the scarcity of labelled datasets for Arabic language increases the challenges of developing new solutions that depend on supervised learning models.

This work presents a model that employs a hybrid DL network to recognize multiple Arabic fonts types including fonts that mimic Arabic handwritten scripts. The hybrid DL model uses a Convolution Neural Network (CNN) and a Bi-Directional Short Long Term Memory network (BDLSTM) and it operates in an end-to-end fashion without the need for character segmentation.

To test the performance of the model, a number of datasets made of (18) different fonts types were compiled. The fonts of the datasets were collected from online sources and they were selected because they exhibit high cursive nature that mimics Arabic handwriting styles. The sample words for generating the custom datasets were extracted from Arabic Wikipedia Dump and they comprise over two million words samples.

Several experiments were performed to examine the model's performance and to assess its generalization capabilities. Despite being a moderate model in terms of its complexity (i.e. can be trained on a single CPU), the same single model was able to achieved (98.76%) in Character Recognition Rate (CRR) and (90.22%) in Words Recognition Rate (WRR) for all the tested font types. However, the model demonstrated degradation in its performance when it was tested on an unseen dataset or noisy images. In the case of unseen dataset, it achieved a CRR success rate of (85.15%), while in the case of noisy images it achieved a CRR success rate of (77.29%). However, resolving these issues require additional investigations, which are out of the scope of this work.

The remaining of the paper is organized as follows: In Section II the challenges related to Arabic text recognition are discussed. In Section III an overview of the related work is presented. Section IV presents the compiled dataset. The proposed model and its process flow is presented in Section V. Section VI presents the experiments and the obtained results and finally, Section VII concludes the paper and identify some future research avenues.

## II. CHALLENGES RELATED TO ARABIC TEXT RECOGNITION

Arabic writing system is used by different nations around the world, this includes the (21) Arab countries as well as other nations such as Kurdish, Pashto, Persian, Sindhi, Uighur and Urdu. The base writing system is made of the base alphabets of Arabic language, which consist of (28) alphabets and (10) Hindi format numerals. The alphabets are written from right-to-left while the numbers are written from left-to-right. On the other hand, the shape of Arabic alphabets can change according to their position in the word. Fig. 1 below shows the based alphabets of Arabic language as well as their variations according to their position in the word.

A main challenge of Arabic writing system is related to its cursive nature knowing that alphabets are written in a joint flowing style. In this respect, characters in Arabic script, whether handwritten, typed or printed are connected within words and they might overlap within the same word or across words (i.e. inter and intra overlapping). In addition, spaces can occur within words and across them while various Arabic characters share the same main shape (e.g. ba, ta, tha as shown in Fig. 1). However, these characters are distinguished by the number of dots added under or above the base alphabet, which increase the challenge of identifying the correct alphabet.

Further, the shapes of Arabic characters are represented using different glyphs according to the characters' position in the word. Accordingly, different shapes are used when the character appears at the beginning, middle or at the end of a word.

| pronounce | end | mid | begin | isolated | pronounce | end | mid | begin | isolated |
|---|---|---|---|---|---|---|---|---|---|
| ḍād | ض | ضـ | ضـ | ض | ʾalif | ـا | | | ا |
| ṭāʾ | ط | ـطـ | طـ | ط | bāʾ | ـب | ـبـ | بـ | ب |
| ẓāʾ | ظ | ـظـ | ظـ | ظ | tāʾ | ـت | ـتـ | تـ | ت |
| ʿayn | ع | ـعـ | عـ | ع | thāʾ | ـث | ـثـ | ثـ | ث |
| ghayn | غ | ـغـ | غـ | غ | jīm | ـج | ـجـ | جـ | ج |
| fāʾ | ف | ـفـ | فـ | ف | ḥāʾ | ـح | ـحـ | حـ | ح |
| qāf | ق | ـقـ | قـ | ق | khāʾ | ـخ | ـخـ | خـ | خ |
| kāf | ك | ـكـ | كـ | ك | dāl | ـد | | | د |
| lām | ل | ـلـ | لـ | ل | dhāl | ـذ | | | ذ |
| mīm | م | ـمـ | مـ | م | rāʾ | ـر | | | ر |
| nūn | ن | ـنـ | نـ | ن | zāy / zayn | ـز | | | ز |
| hāʾ | ه | ـهـ | هـ | ه | sīn | ـس | ـسـ | سـ | س |
| wāw | و | | | و | shīn | ـش | ـشـ | شـ | ش |
| yāʾ | ي | ـيـ | يـ | ي | ṣād | ـص | ـصـ | صـ | ص |

Fig 1.    Arabic base Alphabet.

Similarly, the shape of Arabic characters in printed text varies depending on the used font, as well as its formatting and printing style. Additionally, natural languages that use the Arabic writing system extends the base alphabets by adding special diacritics over some characters to better adapt the writing system to the phonemes of the designated language. A thorough discussing about these challenges can be found in [1]. All these characteristics make the recognition of Arabic text a challenging task, especially for the models that depend on segmenting characters prior to the recognition process [2].

The next section presents some of the related work that was introduced to address some of these challenges and the approaches that were followed.

## III. RELATED WORK

The recognition of Arabic text is still a challenging task because of many intricate features related to the nature of Arabic writing system [3]. Work in this domain is an active research area where many models are continuously proposed for the problem of automatically recognizing printed or handwritten text. Each of these domains has its own challenges and requirements. The challenges of recognizing printed Arabic text are driven by the need for a ubiquitous model that can efficiently recognize Arabic text that is printed using multiple font types and using different formatting styles. On the other hand, the challenges facing the Arabic handwritten text are driven by its high variety due to the diversity of individuals writing styles. In this section, we present an overview about the related work in both domains and the methods that were employed to recognize text.

A recent model for recognizing printed Arabic characters in isolation mode was presented in [4] which applied K-Nearest Neighbor (KNN) and Random Forest Tree (RFT) algorithms to recognize Arabic text. That model used statistical methods to extract features from text images. These features included the dimensions of the text shape, the transition of pixels, the number of black vs white pixels and regional ratios of pixel values. The KNN classifier achieved a successful rate of (98%), while the RFT classifier achieved (87%). Similarly, the authors in [3] introduced a model for recognizing Arabic printed text using linear and nonlinear regression. In that work, text in images were initially thinned and segmented into sub words. Next, the relations between word segments were represented using a numerical coding scheme that represented characters as a sequence of points, lines, ellipses and curves. Using that scheme, a unique code was established for each character form and a unique list of codes were used to recognize each font type. Finally, linear regression technique was used to validate the representations against a ground truth table using distance measures. The model was evaluated using (14000) words samples and it has achieved an accuracy rate of (86%).

In [5], the authors proposed a model for segmenting Arabic printed text that can serve as a preliminary step in the text recognition process. The model that was presented in that work applied contours analysis and template matching techniques to recognize text. The contour segmentation was determined by the local minima values of the contour and the template-based technique involved scanning the positions of black pixels after segmenting the text into lines and sub words using horizontal and vertical projections. The model was evaluated using a custom multi-font corpus and was also benchmarked against five other methods. The model achieved an enhanced accuracy over the other models with a score of (94.74%).

Arabic text recognition research was also influenced by the progress that was achieved in deep learning technology. Earliest work in implementing DL approaches to Arabic text recognition can be traced back to [6]. In that work, a Multi-Dimensional Long Short Memory (MDLSTM) network and Connectionist Temporal Loss (CTC) were used to recognize Arabic text in images. The model was tested on the IFN/ENIT dataset of Tunisian handwritten Town names [7] and an accuracy levels of (91.4%) were reported.

A more recent work in the field of Arabic text recognition using DL models was carried out by [8]. The domain of that work was the recognition of Arabic script in historical Islamic manuscripts. The presented model used various preprocessing techniques to enhance the quality of the scanned images and to segment the text prior to the recognition process. CNNs were used to recognize the preprocessed text and accuracy levels ranging from (74.29%) to (88.20%) were reported.

In [9], a model for recognizing Arabic handwritten text using neural networks was presented. Initially, the noise in images was reduced using multiple image preprocessing techniques. The characters in words were segmented into regions using a threshold-based method and these regions were used to construct feature vectors. The model was examined on a custom dataset collected from volunteered writers and a CRR of (83%) was reported. Similarly, the work of [10] presented a three-layers CNN model for recognizing Arabic handwritten characters in isolation mode. The model was examined on AIA9k [11] and AHCD [12] datasets and CRR of (97.6%) was reported. In [2], the authors presented a DL based model for recognizing Arabic handwritten text using a MDLSTM network and CTC loss function. The objective of that work was to assess the effects of extending the dataset using data augmentation techniques and to compare performance of the extended model against other similar models. The KHAT handwritten dataset [13] was used to train and evaluate the model and a CRR level of (80.02%) was reported. Finally, in [14] a hybrid DL model for detecting printed Urdu text in scanned documents was discussed. The model employed a hybrid combination of CNN and BDLSTM along with CTC loss and it was tested on URDU and APTI datasets [15]. The model was able to achieve CRR rates of (89.84%) and (98.80%), respectively.

Reviewing the related work revealed that there is a shortage in work that addresses Arabic printed text using DL models. Our work should present some footsteps in this research area and provide toolkits that can be utilized to further extend and enhance the achieved outcomes.

## IV. THE COMPILED DATASET

During the last period, several Arabic printed datasets were introduced by the community including: DARPA, APTI, PATDB, APTID/MF, and RCATSS [5], [15]–[18]. Nevertheless, there is no consensus on a standard dataset that can adopted by the community that can be used for benchmarking printed text recognition. Consequently, the available datasets vary in their content types, sizes, formatting styles and fonts types [19]. A more thorough listing of similar datasets can be found in [20].

As stated earlier, the main objective of this work was to examine the performance of DL based models in recognizing Arabic text that was printed using fonts that mimic Arabic hand writing styles. For that purpose, no suitable dataset was found and consequently a number of custom datasets were compiled to serve the purpose i.e. Arabic Multi-Fonts Dataset (AMFDS). These dataset were prepared using the (18) fonts depicted in Fig. 2 below.

In this respect, a custom toolkit for generating the datasets was prepared. This toolkit can be used to generate any number of text image samples using any required font type. It can also be configured to generate samples as separate image files or as a single binary repository for all the samples.

Table I next shows the main characteristics of each generated dataset. As presented in the table, the (ae-Nice) font was selected to generate the single-font datasets. This font type was selected because its printing style clearly exhibits cursive structures that mimic Arabic hand writing script. Similarly, the (ae-Nice) and the (K-Karman) font's types were selected to generate the two-font's datasets. Finally, datasets (4) and (5) in Table I were generated using the font types that are presented in Fig. 2.

Datasets (1, 2, 3) in Table I include duplicate samples because the same set of words was used to generate samples for each font type. In addition, these datasets have minor redundancies within the samples of each font because words were randomly sampled from Arabic Wikipedia Dump and no filtering was applied. Datasets (4 and 5) in Table I are unique (disjoint) datasets where no single word is replicated across the entire dataset. The current version of the datasets contains words samples that have a length of (7 to 10) characters and all the words were generated using font size (26) and bold formatting style.

| Font Type | Font sample |
|---|---|
| ae_Nice | التعرف على النصوص العربية باستخدام تقنيات التعلم الآلي. |
| Aref_Menna | التعرف على النصوص العربية المزيفة باستخدام تقنيات التعلم الآلي. |
| bader_hm-nt | التعرف على النصوص الخطية العربية باستخدام تقنيات التعلم الآلي. |
| Baum Mansb Bln Light | التعرف على النصوص العربية باستخدام تقنيات التعلم الآلي. |
| B Chafenh-th | التعرف على النصوص العربية باستخدام تقنيات التعلم الآلي. |
| B Fantah | التعرف على النصوص العربية المزيفة باستخدام تقنيات التعلم الآلي. |
| BSetash | التعرف على النصوص العربية باستخدام تقنيات التعلم الآلي. |
| B Ziba | التعرف على النصوص العربية باستخدام تقنيات التعلم الآلي. |
| Dast Nevis | التعرف على النصوص العربية باستخدام تقنيات التعلم الآلي. |
| Dava Font | التعرف على النصوص العربية باستخدام تقنيات التعلم الآلي. |
| DroiType Thabth | على النصوص العربية باستخدام تقنيات التعلم الآلي. |
| Ghalam-1 | التعرف على النصوص العربية باستخدام تقنيات التعلم الآلي. |
| Ghalam-2 | التعرف على النصوص العربية باستخدام تقنيات التعلم الآلي. |
| K-Kamran | التعرف على النصوص العربية باستخدام تقنيات التعلم الآلي. |
| K Tabassom | التعرف على النصوص العربية باستخدام تقنيات التعلم الآلي. |
| pH elit Khotdar | التعرف على النصوص العربية باستخدام تقنيات التعلم الآلي. |
| Thanraiik nm_nn Modern | التعرف على النصوص العربية باستخدام تقنيات التعلم الآلي. |
| ToyonAlgntah | التعرف على النصوص العربية باستخدام تقنيات التعلم الآلي. |

Fig. 2. The Set of the Selected Fonts.

TABLE I.    THE PREPARED DATASETS

| Dataset # | Number of fonts | Number of samples | Duplicates in samples | Dataset size | Fonts |
|---|---|---|---|---|---|
| 1 | 1 | 60,000 | Exist | 148MB | ae-Nice |
| 2 | 1 | 120,000 | Exist | 295MB | ae-Nice |
| 3 | 2 | 240,000 | Exist | 490MB | ae-Nice, K-Karman |
| 4 | 18 | 2,160,000 | Exist | 6200 MB | The fonts shown in Fig. 2 |
| 5 | 18 | 450,000 | Unique words | 1300MB | The fonts shown in Fig. 2 |

Further, each dataset is comprised of two main data files: a labels file and binary file. The labels file is a normal text file that contains details about the word samples, this includes the Arabic word represented by the image, the font type, the font style, the font size and a value that represent the starting index of that image in the binary file. Hence, the byte stream of the designated image begins at the starting index and spans to a length equals to the image's size (in bytes). This binary file represents a single repository for all the images in the dataset. Unlike the common adopted approaches of using single image file for each word sample, the format presented in this work is more appropriate for addressing large data files with large number of samples and it is more scalable as it facilitates moving datasets around different execution environments i.e. cloud based environments, it also facilitates the processing of image data in terms of loading, preprocessing and training.

The datasets that were used in the experiments are made publically available at [1], similarly, the toolkit that can be used to generate different samples is made publically available at [2].

## V.    PROPOSED MODEL

In general, text recognition systems implement a series of tasks before recognizing text in images. These tasks can be classified into five main categories which includes: the normalizing of document images to enhance their quality, the detection of text regions within a document and segmenting text accordingly, the extraction of useful features from text, implementing the recognition process and employing post processing techniques to enhance the accuracy of the achieved results. The focus of this work is on the recognition process; while the other tasks are out of the current scope and might be addressed in future research.

The design and the implementation of the proposed model was based on the work presented in [21]. In that work, a hybrid NN for recognizing handwritten text in scanned historical documents was presented. The model presented in this work employs the same intuition and adapts the model to recognize different styles of Arabic printed text.

The proposed model is comprised of two main components; a Convolutional NN and a Recurrent NN. These networks are stacked together in an end-to-end fashion that can perform word-level recognition without the need for character level segmentation.

TABLE II.    MODEL DESIGN

| CNN | | | | |
|---|---|---|---|---|
| Layer | Filter size | # of filters | Pooling window | Output size |
| 1 | (5, 5) | 32 | (2, 2) | (64,16,32) |
| 2 | (5, 5) | 64 | (2, 2) | (32,8,64) |
| 3 | (3, 3) | 128 | (1, 2) | (32,4,128) |
| 4 | (3, 3) | 128 | (1, 2) | (32,2,128) |
| 5 | (3, 3) | 256 | (1, 2) | (32,1,256) |
| BDLSTM | | | | |
| Layer | # of hidden units | | | |
| 1 | 256 x 2 (forward and backward) | | | |
| 2 | 256 x 2 (forward and backward) | | | |

Table II presents the specifications of the proposed model.

As shown in the previous figure, the filter sizes in the initial two layers of the CNN employs filters of size (5, 5) units. This filter size is suitable for extending the receptive field of the early layers of the network. The three remaining convolution layers in the network employed filters sizes of (3, 3).

Further, the convolution process in the model employed zero padding so that it can preserve the size of the input image throughout the convolution process. The pooling process in the initial two layers used a sliding window of size (2x2) while the remaining three layers used a window of size (1, 2).

The CNN is stacked on top of a BDLSTM in an end-to-end manner. The BDLSTM had (2) LSTM layers, each layer had two LSTM cells that implements the forward and backward passes of inputs in the network, and each LSTM cell had (256) hidden units. Fig. 3 next shows the general architecture of the model.

The processing of the model starts with the CNN accepting input images of size (128×32). Therefore, prior to injecting the images into the model, these images were resized to a size of (128×32) units. The resizing process changes the shape of the input images by compressing it and shifting the location of the text. However, CNNs are shift invariant and they are tolerable to such variances. Fig. 4 below shows the effects of applying the resizing process on a sample word.

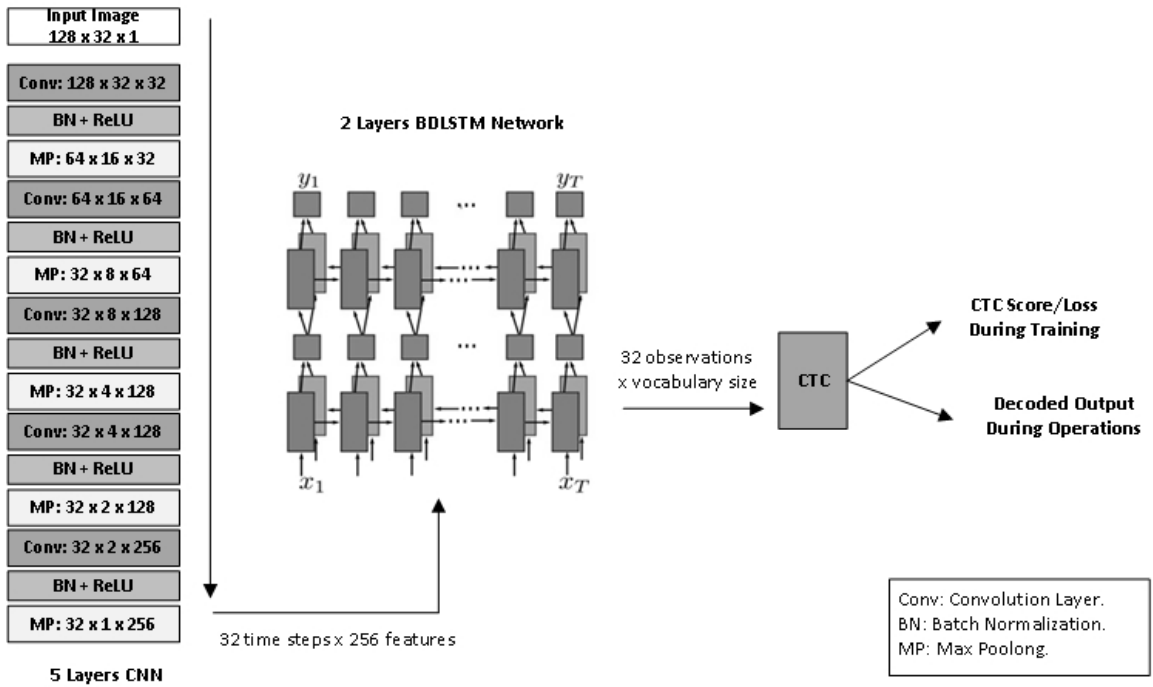

Fig 3.    General Architecture of the Proposed Model, which is Comprised of (5) Layers CNN, (2) Layers BDLSTM and CTC Loss Function.

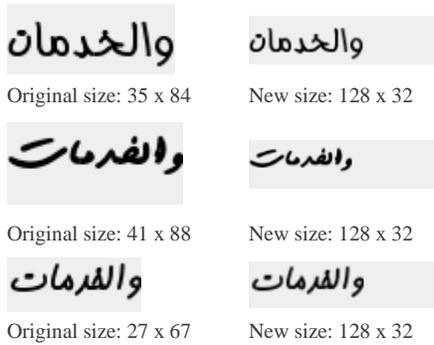

Fig 4.    Input Images before and after Resizing.

Once an input image is received by the model, the CNN implements four main operations: Convolution, Batch Normalization, Activation and Max Pooling.

The convolution process extracts the relevant features from the input image through the filters and passes the values to a batch normalization process to mitigate the effects of covariance shift [22]. Next, ReLU activation function is implemented to eliminate negative values and to minimize the effects of vanishing gradients. The results of the activation function are passed to a max-pooling layer which performs sub sampling by selecting the most relevant features and gradually downsizes the input size into an array of (32 time-steps×256 features). Fig. 8(a-f) in Appendix (A) show the output of the convolution layers for the sample Arabic word (al rasmeya "الرسمية"). These figures demonstrate how the earlier layers in the CNN captures detailed features while the later layers capture more generalized features. In addition, Fig. 8(f) in the appendix shows the output of the final layer of the CNN which demonstrate how the features are concentrated in the lower indexes (0-10 of 32) of the output array as the words length in the experiments ranged between (7-10) characters.

The final output of the CNN is passed to the BDLSTM network, which learns the sequence or the temporal dimension in the input image. The output of the BDLSTM network is an array of size (32-observations×vocabulary size) knowing that a single character in the input text might be represented by one or more observation sequence.

Sample outputs of the BDLSTM network for Arabic word (al rasmeya "الرسمية") are presented in Fig. 8(h-j) in Appendix (A). The figures demonstrate how the values for a specific output sequence increased at locations that corresponds to its index in the vocabulary that shown in Fig. 5. This vocabulary represents the unique characters that are represented in the custom dataset.

Next, the BDLSTM observations are passed to a CTC loss function, which performs a probabilistic based mapping between the observations and ground-truth labels. According to the model's specifications, the CTC function is able to recognize words of size (32) characters while each time step sequence can represent one of the (38) different characters that are shown in Fig. 5.

Finally, the CTC function computes the loss value and back propagates it to the network to initiate the end-to-end learning process. RMSOptimizer [23] was used to implement the optimization process in the proposed model.

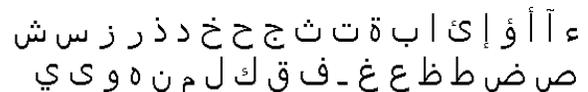

Fig 5.    List of Unique Characters in the Compiled Dataset.

## VI. EXPERIMENTS AND RESULTS

To train and examine the proposed model, a TensorFlow based implementation was prepared using Python, this implementation is made publicly available at [3].

The model was trained and tested using Google Colab platform. That platform provides computing environments that includes hardware acceleration that can be used to train different DL models. The environment that were used for implementing the model in this work had the following specifications:

- Python version: 3.6.
- Tensor flow version: 2.0.
- Hardware Accelerator: 12GB NVIDIA Tesla K80 GPU.
- RAM: 12.72 GB.
- HDD: 64.4 GB.

To evaluate the model, CRR and Words Recognition Rate (WRR) measures were employed, the formulas of these measures are shown below:

$$CRR = \frac{\sum Levenshtein\ EditDistance(Recognized\ Text, Ground\ Truth)}{Total\ Number\ of\ Processed\ Characters}$$

$$WRR = \frac{Number\ of\ correctly\ identified\ words}{Total\ number\ of\ words}$$

The model was examined through a number of testing scenarios using the five datasets that were presented in Table I. For this purpose, the datasets were split into a (train, validate, and test) segments using a ratio of (80%, 10%, and 10%) respectively. Several testing scenarios were implemented to examine the model's performance and the results of these testing scenarios are shown in Table III below.

Initially, the model was tested using a relatively small single-font dataset (i.e. dataset #1 in Table I). The results of this test were (97.5%) for CRR and (85.18%) for WRR. Next, the model was tested on a larger dataset with the same font setting (dataset #2 in Table I) and the CRR success rate enhanced to (99.04%) while the WRR achieved (94.29%). In general, the model achieved good results when it was examined on a single-font type dataset regardless of the type or the formatting style of the tested font.

To examine the model's performance on a more diverse dataset, it was tested using a two-fonts dataset (dataset #3 in Table I). Using this dataset set, the model achieved a CRR rate of (99.88%) and a WRR rate of (99.2%).

The high success rates that were achieved in the previous testes were further challenged by testing the model using an extended dataset that is comprised of two million word samples (i.e. dataset #4 in Table I). In this testing scenario, the model achieved good results in CRR (99.27%) but it demonstrated a minor degradation in WRR (94.32%). This behavior can be justified by emphasizing that the WRR is highly dependent on the accuracy of the CRR, and a minor flaw in the recognition of a single character shall affect the recognition of all the related words, especially that the dataset samples included relatively long words (a length of 7–10 characters).

To evaluate the model more accurately in terms of overfitting, it was trained on a new dataset that has no duplication (dataset #5 in Table I). Using this disjoint dataset, the performance of the model demonstrated a minor degradation in the reported accuracy, but it still achieved good results which were (98.76%) for CRR and (90.22%) for WRR.

To test the model's generalization capabilities, an experiment was performed using a new dataset with words length of (5-6) characters. Although the model was not trained on this length of words, it was able to achieve a CRR success rate of (98.71%) and a WRR success rate of (92.4%). Obviously, this was an indication that the model can be generalized beyond the samples that were used in the training process.

In the same aspect, a pilot testing for the model was implemented on the external APTI database [15]. In this experiment, an Arabic typesetting font of size (10) was selected. Although the model was not trained or fine-tuned on this database, it was able to achieve a CRR success rate of (85.15%), while the WRR success rate was degraded to (23.7%). Again, this performance degradation can be justified by the high correlation between the WRR and the accuracy levels of CRR.

Finally, the behavior of the model was tested on some noisy datasets. For this purpose, a selected set of test samples were induced with salt and pepper (S&P) noise and Speckle noise. Fig. 6 below shows a sample word before and after noise transformations.

For this last experiment, the model achieved an acceptable CRR success rate of (82.01%) for the induced S&P noise, while it achieved (77.29%) for both S&P and Speckle noises. However, for the WRR success rates, the model reported (21.48%) for S&P alone and (14.18%) for both noises; S&P and Speckle. Fig. 7 below summarizes the model's performance in all conducted experiments.

The conducted experiments show that the model was able to achieve high accuracy results when it was trained on a specific set of fonts. This might be an indication that the number of the supported fonts could be extended using the same intuitions in a more complex model i.e. larger number of parameters. In addition, several techniques can be incorporated to enhance the WRR accuracy level that has degraded under some testing scenarios. This might include employing post-processing techniques such as language models in order to enhance the overall accuracy of the WRR. In addition, various image-preprocessing techniques can be examined to mitigate the effects of noisy environments.

---

TABLE III. PERFORMANCE RESULTS OF THE MODE

| # | Experiment Name | Dataset | Validation Accuracy % | | Test Accuracy % | |
|---|---|---|---|---|---|---|
| | | | CRR | WRR | CRR | WRR |
| 1 | Single font model | 1 | 98.37 | 90.28 | 97.5 | 85.18 |
| 2 | Single font model – larger dataset | 2 | 99.148 | 94.66 | 99.044 | 94.29 |
| 3 | Two fonts model | 3 | 99.93 | 99.5 | 99.88 | 99.2 |
| 4 | (18) fonts, duplicate words across fonts types | 4 | 99.38 | 94.84 | 99.27 | 94.32 |
| 5 | (18) fonts, unique words across the dataset | 5 | 98.81 | 90.53 | 98.76 | 90.22 |
| 6 | Testing model generated in expriment 5 above on five character words | - | - | - | 98.71 | 92.4 |
| 7 | Testing model generated in expriment 5 above on APTI dataset – new font | - | - | - | 85.15 | 23.7 |
| 8 | Testing model generated in expriment 5 above with salt & pepper noise. | - | - | - | 82.01 | 21.48 |
| 9 | Testing model generated in expriment 5 above with salt & pepper and speckle noise. | - | - | - | 77.29 | 14.18 |

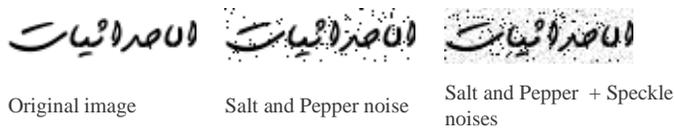

Original image    Salt and Pepper noise    Salt and Pepper + Speckle noises

Fig 6. Sample Images with Noise Adjustments.

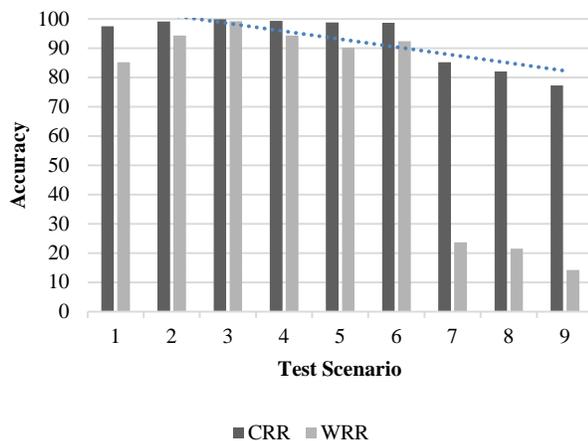

Fig 7. Model Performance for All Conducted Experiments.

## VII. CONCLUSION

In this work, a hybrid DL model for recognizing Arabic text in images was presented. The objective of this work was to examine the models competency in recognizing Arabic text that was printed using multiple font types, including fonts that exhibit high cursive nature that mimic Arabic handwriting script. The proposed model demonstrated good CRR in most testing scenarios including the testing on a disjoint dataset, the testing on a pilot external database and the testing under noisy environments. The overall performance of the model is open for more enhancements through incorporating language models to enhance the overall WRR accuracy as well as using image processing techniques to mitigate the effects of noise in images. The same model can also be examined in recognizing Arabic handwritten text. Such measures might be investigated in a future work.

APPENDIX A

*A. CNN and RNN Layers Sample Outputs*

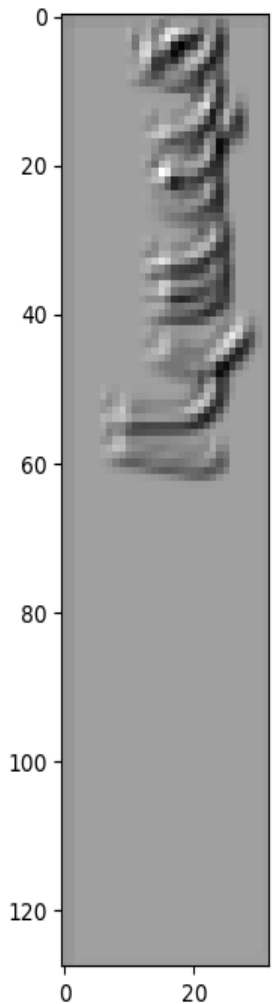 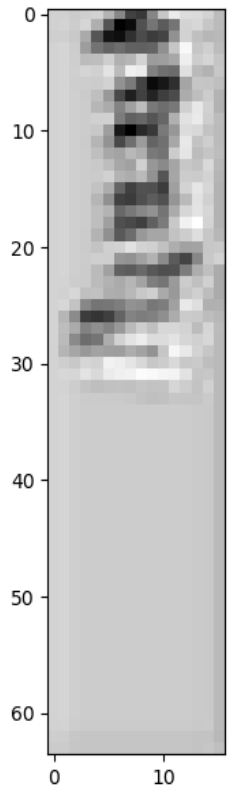 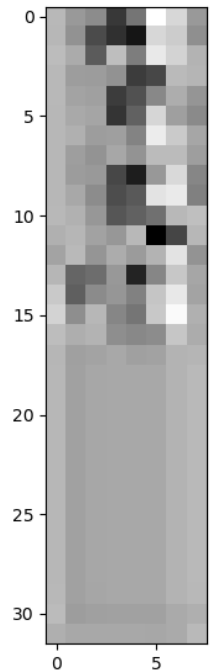 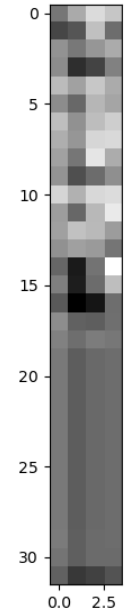 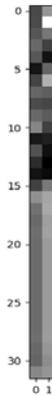

(a) CNN Layer [1,32].    (b) CNN Layer [2,32].    (c) CNN Layer [3,32].    (d) CCN Layer [4,32].    (e) CCN Layer [5,32].

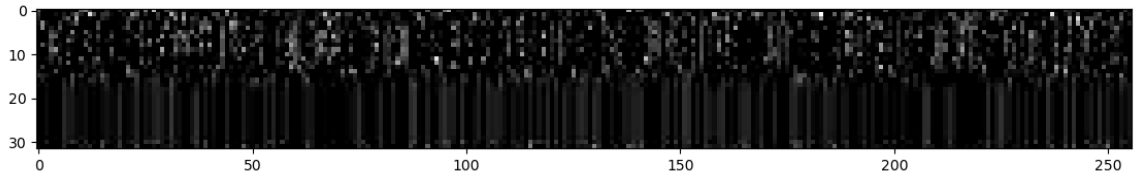

f: Final Output of the CNN (32 Time Steps × 256 Features).

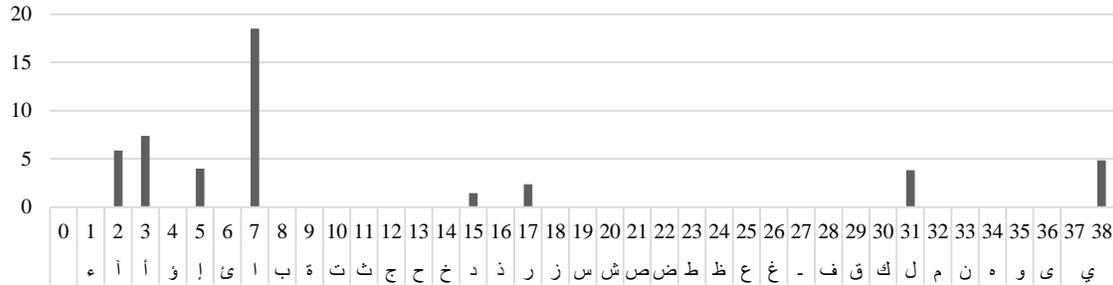

g: Time step [1 of 32] – Highest Observation at Vocabulary Position [7] = Character Alef.

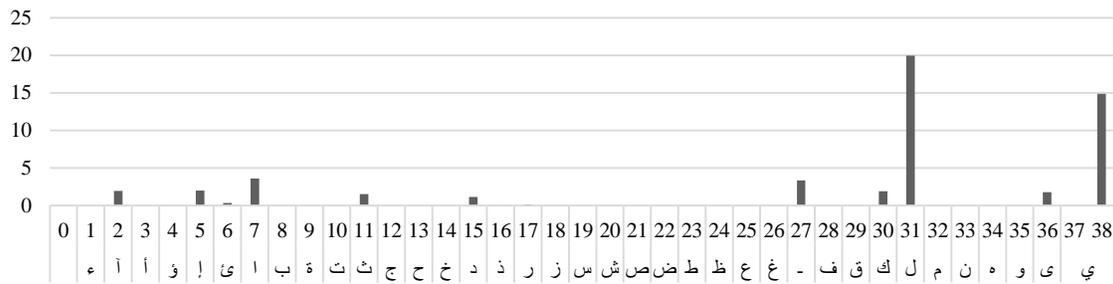

h: Time step [2 of 32] – Highest Observation at Vocabulary Position [31] = Character Laam.

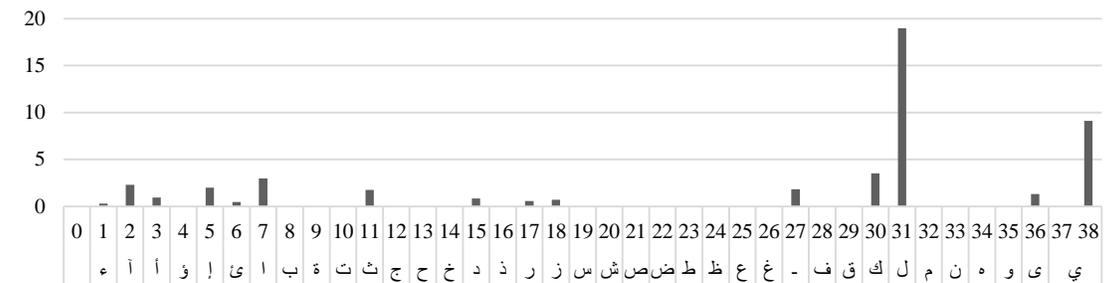

i: Time Step [3 of 32] – Highest Observation at Vocabulary Position [31] = Also Character Laam.

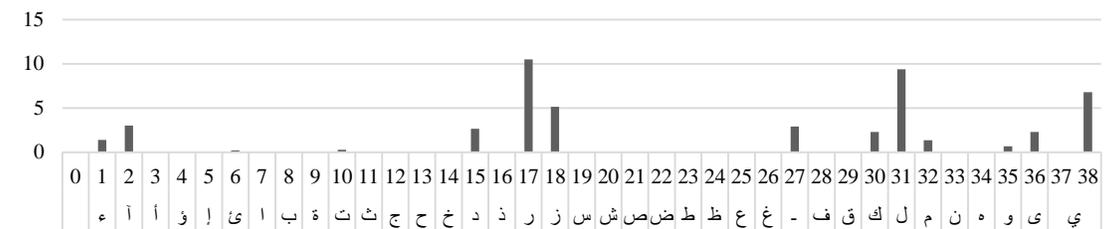

j: Time Step [4 of 32] – Highest Observation at Vocabulary Position [17] = Alphabet Raa, also High Value at Position [31] Character Laam.

Fig 8. (a-j) CNN and RNN Layers Sample Outputs.